\documentclass[letterpaper, 10 pt, conference]{ieeeconf}  % Comment this line out if you need a4paper

\IEEEoverridecommandlockouts                              % This command is only needed if 
\usepackage{amsmath,amsfonts}
\usepackage{algorithmic}
\usepackage{algorithm}
\usepackage{array}
\usepackage[caption=false,font=normalsize,labelfont=sf,textfont=sf]{subfig}
\usepackage{textcomp}
\usepackage{stfloats}
\usepackage{url}
\usepackage{verbatim}
\usepackage{graphicx}
\usepackage{cite}
\usepackage{tabularray}
\usepackage{multirow}
\usepackage{xcolor}
\usepackage{tabularx}
\usepackage{booktabs}
\usepackage{graphicx}
\usepackage{caption}
\usepackage{setspace}
\usepackage{hyperref}

\title{\LARGE \bf
PolyFit: A Peg-in-hole Assembly Framework for Unseen Polygon Shapes via Sim-to-real Adaptation 
}

\author{Geonhyup Lee*, Joosoon Lee*, Sangjun Noh, Minhwan Ko, Kangmin Kim, and Kyoobin Lee†% <-this % stops a space
\thanks{*Equal Contribution}%
\thanks{All authors are with the School of Integrated Technology (SIT), Gwangju Institute of Science and Technology (GIST), Cheomdan-gwagiro 123, Buk-gu, Gwangju 61005, Republic of Korea. 
† Corresponding author: Kyoobin Lee {\tt\small kyoobinlee@gist.ac.kr}}%
}
\begin{document}

\maketitle
\thispagestyle{empty}
\pagestyle{empty}

%%%%%%%%%%%%%%%%%%%%%%%%%%%%%%%%%%%%%%%%%%%%%%%%%%%%%%%%%%%%%%%%%%%%%%%%%%%%%%%%
\begin{abstract}

The study addresses the foundational and challenging task of peg-in-hole assembly in robotics, where misalignments caused by sensor inaccuracies and mechanical errors often result in insertion failures or jamming. This research introduces PolyFit, representing a paradigm shift by transitioning from a reinforcement learning approach to a supervised learning methodology. PolyFit is a Force/Torque (F/T)-based supervised learning framework designed for 5-DoF peg-in-hole assembly. It utilizes F/T data for accurate extrinsic pose estimation and adjusts the peg pose to rectify misalignments. Extensive training in a simulated environment involves a dataset encompassing a diverse range of peg-hole shapes, extrinsic poses, and their corresponding contact F/T readings. To enhance extrinsic pose estimation, a multi-point contact strategy is integrated into the model input, recognizing that identical F/T readings can indicate different poses. The study proposes a sim-to-real adaptation method for real-world application, using a sim-real paired dataset to enable effective generalization to complex and unseen polygon shapes. PolyFit achieves impressive peg-in-hole success rates of 97.3\% and 96.3\% for seen and unseen shapes in simulations, respectively. Real-world evaluations further demonstrate substantial success rates of 86.7\% and 85.0\%, highlighting the robustness and adaptability of the proposed method. Videos of data generation and experiments are available online at \href{https://sites.google.com/view/polyfit-peginhole}{https://sites.google.com/view/polyfit-peginhole}.

\end{abstract}

\section{Introduction}
\label{sec:Introduction}

\begin{figure}[ht!]
\centering
\includegraphics[width=\linewidth]{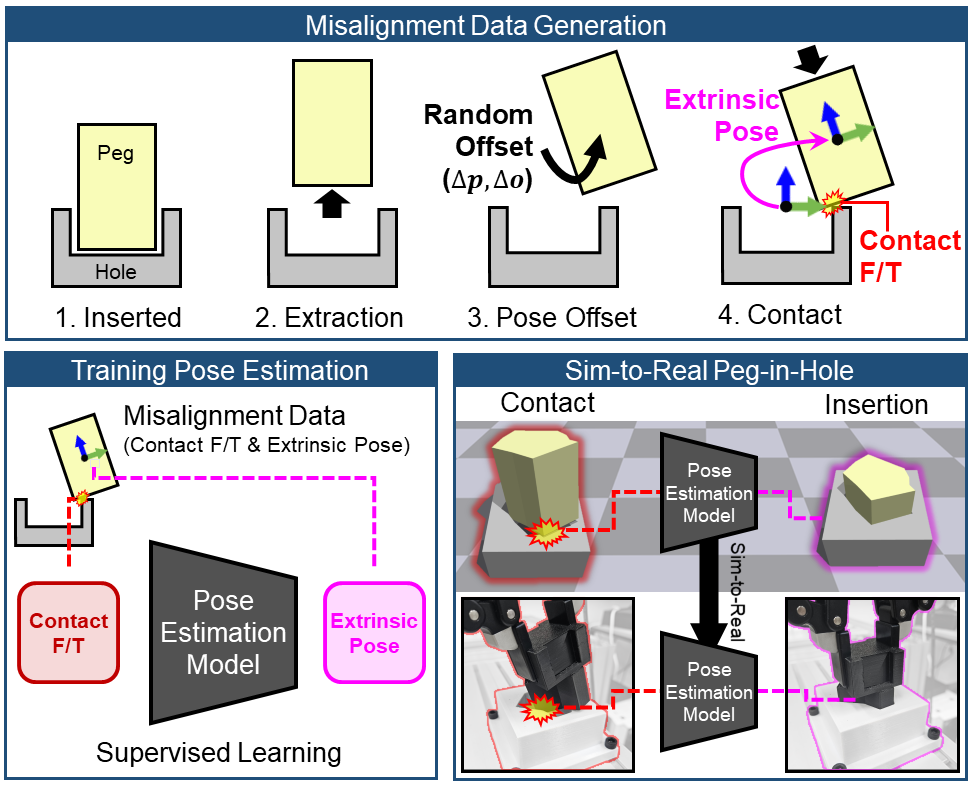}
\caption{Schematic illustration of addressing misalignment using the proposed PolyFit framework for a peg-in-hole task.}
\label{fig:figure_1}
\end{figure}

Peg-in-hole assembly using robotics is a fundamental yet challenging task. Misalignment issues often occur between assembled parts due to sensor inaccuracies and mechanical errors, potentially resulting in system damages like insertion failures or jamming. Various methods based on compliance control and heuristic algorithms have been proposed to address misalignment in peg-in-hole assembly \cite{bruyninckx1995peg, tang2016autonomous, park2020compliant, van2018comparative}. However, traditional methods may require manual calibration for specific peg-hole shapes or dynamic modeling of contact situations. These requirements complicate the development of assembly strategies for different component shapes and limit generalization across shapes. The proposed PolyFit framework for the peg-in-hole task is illustrated in Fig.\ref{fig:figure_1}.

% --------------------------------------------- Fig.1-----------------------------------------
% \vspace{-5pt}
% --------------------------------------------- Fig.1-----------------------------------------

Learning-based peg-in-hole methodologies have been proposed to address the challenges mentioned above \cite{xie2022learning, ding2019transferable, azulay2022haptic, luo2019reinforcement, hebecker2021towards}. One study guided a peg into a hole by visually minimizing the seam through vision-based reinforcement learning (RL), showcasing adaptability to unseen shapes \cite{xie2022learning}. However, vision-based methods may be sensitive to external conditions such as occlusion, varying lighting conditions, and camera placement.

Meanwhile, methodologies based on Force/Torque (F/T) sensors operate through direct contact with objects and exhibit less sensitivity to external environments challenging for vision sensors. Several studies have utilized contact F/T data sampling in the real world for peg-in-hole tasks, employing learned contact dynamics and an insertion policy to transfer to new shapes and validate against untrained ones \cite{ding2019transferable, azulay2022haptic}. While demonstrating adaptability to unseen objects, collecting real-world data for these studies can be challenging owing to reduced efficiency and the risk of potential damage to the robot.
In response to these challenges, contact-based sim-to-real adaptation has emerged as a promising strategy to leverage simulations for data collection \cite{luo2019reinforcement, hebecker2021towards}. While simulation environments offer a safer and more data-rich context for learning, defining and randomizing dynamic parameters present substantial obstacles to efficient sim-to-real adaptation.

The proposed PolyFit introduces a significant paradigm shift in peg-in-hole assembly approaches, moving from a RL methodology to a supervised learning-based approach. Addressing the limitations of traditional methods, PolyFit is an F/T-based supervised learning framework designed for 5-degrees-of-freedom (DoF) peg-in-hole assembly. The framework undergoes extensive training in simulation and is adapted for real-world environments.

\begin{figure*}[ht!]
\centering
\includegraphics[width=1.0\textwidth]{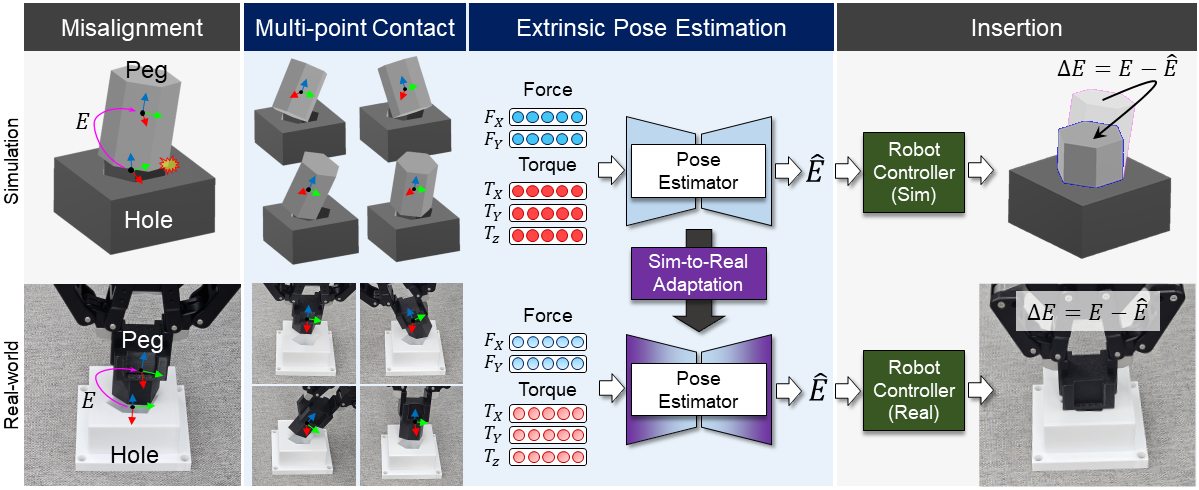}
\caption{Overview of the proposed PolyFit supervised learning-based peg-in-hole strategy. }
% \vspace{-10pt}
\label{fig:figure_2}
\end{figure*}

The simulation dataset includes a diverse array of peg-hole shapes and extrinsic poses between the peg and hole, utilizing the concept of extrinsic poses introduced by Ma et al. \cite{ma2021extrinsic}. This dataset also incorporates corresponding contact F/T readings. PolyFit utilizes this simulation dataset to estimate extrinsic poses using only F/T data, enabling it to adjust the peg pose to correct misalignments. A multi-point contact strategy is implemented to enhance pose estimation accuracy, considering that identical F/T data can occur in different poses.
PolyFit is trained on polygonal shapes with 4, 5, and 6 vertices, and its robustness is validated through tests on more complex, unseen configurations with 7, 8, 9, and 10 vertices, demonstrating its ability to generalize to new configurations. Sim-to-real adaptation methods are introduced to extend the model to the real-world. The adaptation method utilizes a minimal set of sim-real paired data, capturing identical contact F/T and poses in simulated and real environments. This adaptability is validated through peg-in-hole experiments on unseen polygon shapes, confirming the model's precision and effective adaptability to real-world applications. An overview of the PolyFit model is presented in Fig. \ref{fig:figure_2}. Despite relative errors induced by perception issues or noise, PolyFit ensures successful task completion through iterative multi-point contact and F/T data acquisition.

The contributions of this work can be summarized as follows:
\begin{enumerate}
\item Development of a supervised learning framework for 5-DoF peg-in-hole assembly, leveraging only F/T data for extrinsic pose estimation and misalignment correction.\item Introduction of a sim-to-real adaptation method for a pose estimation model, and validation of peg-in-hole assembly in the real-world, demonstrating generalizability across untrained polygon shapes.
\end{enumerate}

% width=0.85

\section{Related Work}
\label{sec:related work}
\subsection{Learning-based Method for Assembly}
{\bf{Vision-based Approach:}}
Significant research has been conducted on vision-based methodologies for robotic peg-in-hole tasks \cite{schoettler2020deep, shi2023sim, haugaard2021fast, shen2023learning, spector2021insertionnet, spector2022insertionnet, nair2023learning}. 
Studies focusing on specialized reward mechanisms, such as sparse rewards and goal-oriented images, have been carried out using RL frameworks for complex industrial insertion problems \cite{schoettler2020deep, shi2023sim}. Furthermore, learning-driven vision-based servo techniques have demonstrated assembly-task improvements \cite{haugaard2021fast, shen2023learning}. However, these methods have been primarily tested on a limited number of pegs and holes.
Therefore, approaches that utilize a "residual policy" have been explored to extend capabilities across a broader array of assembly components and enhance efficiency and flexibility in real components used in robotic assembly \cite{spector2021insertionnet, spector2022insertionnet}. However, these strategies encountered limitations in adapting to new configurations.
Offline RL was employed to reduce the requisite exploration in traditional RL and facilitate the efficient learning of assembly policies across various configurations, with online fine-tuning concurrently explored for swift adaptation to new configurations \cite{nair2023learning}.
However, this methodology encountered limitations owing to the dimensional constraints of the action space that only considered positional errors. As previously noted, vision-based methods face challenges due to sensor noise, occlusion, and variable lighting conditions.
In response to these challenges, the proposed PolyFit framework employs F/T modalities to execute 5-DoF peg-in-hole tasks, thereby offering a robust performance that effectively mitigates the challenges encountered by vision sensors.

{\bf{Force-based Approach:}}
The limitations of vision-based methods have led researchers to explore alternative approaches, particularly those utilizing force-based methods. RL with F/T information has been employed to learn force controller action policies, resulting in a precise insertion strategy \cite{luo2019reinforcement, beltran2020variable}. However, this method faced challenges when dealing with unseen shapes. 
% To overcome this limitation, subsequent research focused on shape generalization in assembly tasks \cite{ding2019transferable, dong2021tactile, zhao2022offline}.
To overcome this limitation, subsequent research focused on shape generalization in assembly tasks \cite{ding2019transferable, zhao2022offline}.
A method was developed using contact data sampling that efficiently learned a dynamic model using multi-pose F/T state representations, enabling quick adaptation to new peg-in-hole configurations with minimal trials \cite{ding2019transferable}. 
Offline meta-RL was introduced to enhance learning across diverse components and accelerate adaptation through online fine-tuning \cite{zhao2022offline}. Despite these advancements, significant challenges persist. The acquisition of real-world data introduces concerns regarding efficiency and safety \cite{luo2019reinforcement, ding2019transferable, zhao2022offline}. Additionally, constraints tied to arbitrarily defined dynamic parameters, such as stiffness and friction \cite{beltran2020variable}, can limit the generalization capabilities of the technique, especially during the transition from simulations to real-world scenarios.
% 우리연구
In response to these challenges, this research proposes a methodology for conducting peg-in-hole tasks based on F/T, bypassing the inefficiencies and safety concerns of real-world data acquisition by employing a dynamic simulator for data collection and learning. Consequently, the proposed method ensures robust performance across different component geometries and conditions without the necessity for extensive data collection in the real world. 

\subsection{Sim-to-Real Adaptation for Assembly}

Adapting from simulation to real-world environments presents a fundamental challenge in robotics owing to inherent discrepancies between the two domains, including differences in physical dynamics and unmodeled factors. Various research initiatives aim to identify and mitigate these disparities, ensuring the effective transfer of learned behaviors from virtual to physical environments \cite{tobin2017domain, peng2018sim, chebotar2019closing, du2021auto}. Domain-randomization techniques, such as visual randomization \cite{xie2022learning, shi2023sim, valassakis2021coarse} and dynamic parameter randomization \cite{luo2019reinforcement, hebecker2021towards}, have demonstrated robustness across a variety of real-world conditions in assembly tasks.
Furthermore, scaled robot force was output continuously as a strategy \cite{scherzinger2019contact}, and scripted actions in simulations and reality have been executed for system identification \cite{kaspar2020sim2real}. A small amount of real-world data and meta-transfer learning has also been utilized \cite{schoettler2020meta}. However, existing methods that were focused on adjusting dynamic parameters between simulations and reality often faced scalability and flexibility issues due to the complexity of these parameters. The proposed method counters these limitations by directly adapting contact dynamics through minimal-paired data, capturing the same contact F/T and pose in sim-real environments, and enabling swift, data-efficient adaptation and robust performance across real-world scenarios.

\section{Problem Description}
\label{sec:problem description}

This study aimed to develop a framework capable of accurately estimating and correcting extrinsic pose misalignments during peg-in-hole assembly tasks using F/T measurements within a supervised learning context. 

While peg and gripper poses are known, the extrinsic pose between the peg and hole remains uncertain, leading to misalignments and subsequent contact during assembly. Force control is applied to stabilize F/T readings for the pose estimation model.
Observing F/T in isolation can yield identical readings for different poses. Therefore, a series of \( m \) rotational actions, referred to as multi-point contact, are executed to acquire comprehensive contact pose information. All F/T values discussed in this paper are obtained from these multi-point contact operations. The model inputs include forces \(F_x, F_y\) and torques \(T_x, T_y, T_z\), excluding the constant z-direction force \( F_z \).

In mathematical terms, F/T is articulated as \(C \in \mathbb{R}^{5 \times k}\), encompassing force \(F \in \mathbb{R}^{2 \times k}\) and torque \(T \in \mathbb{R}^{3 \times k}\), where \(k = m+1\) and \(m\) signify the number of additional contact points derived from multi-point contact. 
The 5-DoF extrinsic pose of the initial contact state, denoted as \(E \in \mathbb{R}^{5 \times 1}\), is estimated without considering the component \(p_z\). It incorporates the relative position \(p = (p_x, p_y) \in \mathbb{R}^{2 \times 1}\) and orientation \(o = (o_x, o_y, o_z) \in \mathbb{R}^{3 \times 1}\) of the peg with respect to the hole. 
The peg-in-hole procedure follows a structured framework that sequentially involves multi-point contact, pose estimation, and peg control based on an estimated extrinsic pose. This framework demonstrates applicability in simulated and real-world environments.

\section{Methodology}
\label{sec:Methodology}

\subsection{Simulation Data Generation}
\begin{figure}[h!]
    \centering
    \includegraphics[width=0.9\linewidth]{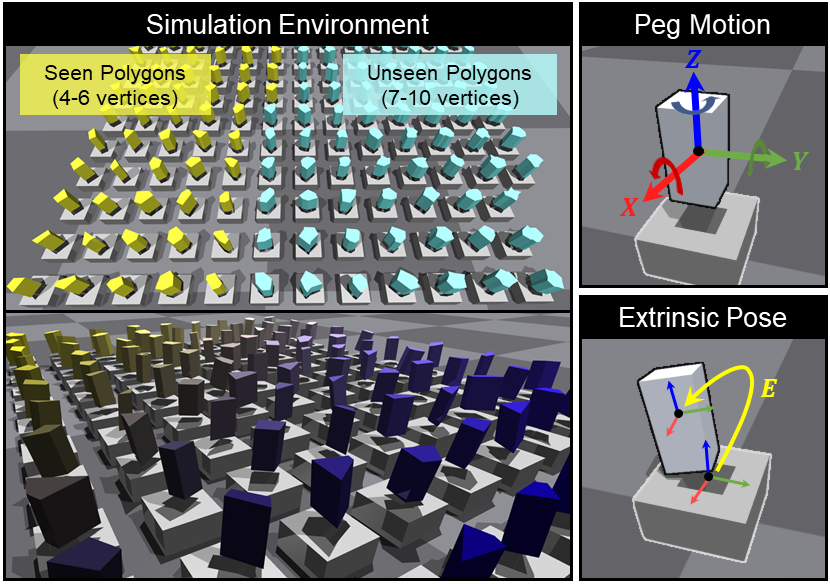}
    \caption{Simulation environment for data generation.}
  \label{fig:figure_3}
\vspace{-10pt}
\end{figure}

{\bf{Simulation environment:}} A simulation environment was established within the Isaac Gym simulator \cite{makoviychuk2021isaac}, as depicted in Fig. \ref{fig:figure_3}. This environment focused predominantly on the peg-hole interaction by intentionally omitting robotic elements. A 6-DoF motion system simulated peg movements, and compliance parameters were adjusted to mirror real-world robotic controllers. It should be noted that these parameters can slightly vary between different real-world implementations. Contact F/Ts were recorded via a centrally located F/T sensor on the peg, with measurements averaged over 50 simulation steps for consistency. Signed Distance Function (SDF)-based collisions with a set resolution of 512 were employed to accommodate the minimal tolerance between the peg and hole \cite{macklin2020local}. Various peg and hole CAD models were developed for seven polygon configurations ranging from 4 vertices to 10 vertices, creating 20 unique shapes per configuration, totaling 140 models.
The polygon geometry was defined by selecting several vertices \( n \) and then randomly assigning radius \( r \) and angle \( \theta \) values, ensuring the total angle sum of 360°\(\left(\sum_{i=1}^n \theta_i = 360^\circ\right)\).
Radius values \(\left\{r_1, \ldots, r_n\right\}\) ranged from 10 to 20 mm. Shapes with angles summing to 180° across three consecutive vertices were excluded to ensure diversity. Furthermore, a tolerance of 1 mm was adhered to for pegs and holes.

{\bf{Peg-hole Misalignment Dataset:}} 
In this study, as illustrated in the misalignment data generation process in Fig. \ref{fig:figure_1}, we generated peg-in-hole misalignment data to train an extrinsic pose estimation model. This process involved extracting a peg from a hole, applying a random pose offset, and subsequently recording the contact F/T and the extrinsic pose at the point of contact.
A random offset sampling strategy introduced random positional \(\Delta p\) and orientational \(\Delta o\) offsets in the peg-hole system to create diverse pose configurations, enabling comprehensive data collection for both extrinsic pose \( E \) and corresponding contact F/T \( C \).
The offset strategy adjusted the initial 5-DoF peg poses with positional offsets \(\Delta p\) within $\pm 10 \text{mm}$ and orientational offsets within $\pm5^\circ$. Subsequently, the PolyFit framework employed a multi-point contact strategy, as outlined in Section \ref{sec:problem description}, involving a sequence of \( m \) additional contacts. In this study, we set \( m \) to 4. Each defined by a controlled $\pm10$° rotation in the x and y orientations relative to the contact direction. This method enhanced the contact F/T data captured in the matrix \( C \), consolidating force and torque measurements from both the initial \( C_0 \) and subsequent \( \{ C_1, \ldots, C_m \} \) contacts. Paired with the extrinsic pose \( E \) of the initial contact, the matrix facilitated the development of a dataset for model training and inference. The dataset included 448 million data points across 140 shapes, each with 3,200 misalignments, distributed as 1,200 for training and 1,000 each for validation and testing. The data generation process was conducted on an RTX 3090 Ti GPU, taking about 3 min per shape.

\subsection{Extrinsic Pose Estimation Network}
\begin{figure}[t]
    \centering
    \includegraphics[width=0.85\linewidth]{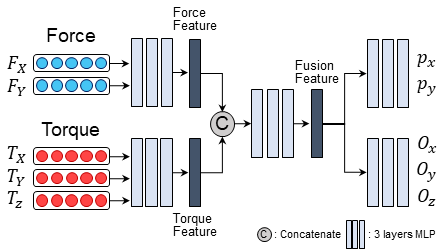}
    \caption{Extrinsic pose estimation network architecture.}
  \label{fig:figure_4}
% \vspace{-10pt}
\end{figure}

The architecture for an effective extrinsic pose estimation network suitable for robotic operations utilized a lightweight design with a multi-layer perceptron (MLP), as shown in Fig. \ref{fig:figure_4}. Inputs comprising force and torque data \( C \), concatenated with values from initial and multi-point contacts, were processed through distinct three-layered MLP encoders, generating separate feature sets for force and torque. These feature sets were then concatenated and passed through a fusion module consisting of a three-layered MLP to produce a unified feature representation. The network's outputs, the extrinsic pose \(\hat{E} \), position, and orientation at the misalignment, were estimated by two separate three-layered MLP heads. Implemented using PyTorch, the network employed the Mean Absolute Error (MAE) as the loss function and the Adam optimizer with an initial learning rate of 0.001. A cosine annealing scheduler was used in 100-epoch cycles for learning rate management during the 300-epoch training. The training, with a batch size of 256, was executed on an RTX 3090 and completed in approximately 30 min.

\subsection{Sim-to-Real Adaptation for Extrinsic Pose Estimation Network}

\begin{figure}[ht]
% \centering
\includegraphics[width=\linewidth]{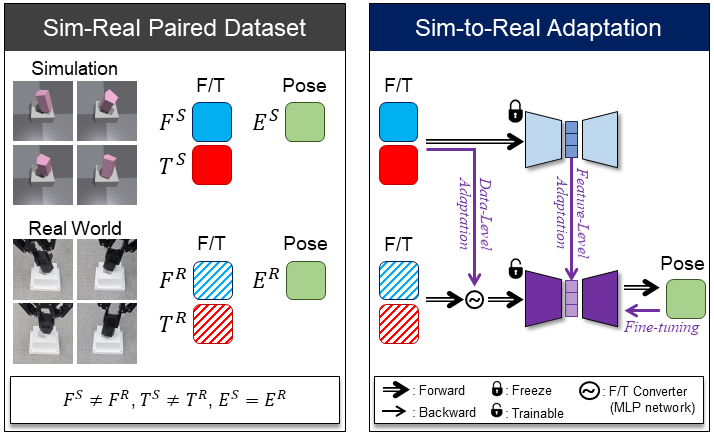}
\caption{Conceptual illustrations of the sim-real paired dataset and sim-to-real adaptation. The contact F/T and poses of the sim-real dataset are utilized in the sim-to-real adaptation, represented with the same color and shape.}
\label{fig:figure_5}
\vspace{-10pt}
\end{figure}

Sim-to-real adaptation is crucial for addressing the performance decline of models trained solely in simulations when applied to real-world tasks. These models often underperform in real scenarios, despite being well-trained in simulations. This issue arises mainly due to the differences in patterns between simulated and real-world data. 
Considering the significant effort and resources required to amass a diverse set of real-world training data, an approach was developed to align real-world contact F/T and features of extrinsic pose estimation models with those from simulations. 
To address this, the proposed approach utilized sim-real paired data capturing contact F/T data with identical shapes and poses in simulated and real environments, as illustrated in Fig \ref{fig:figure_5}.
Proposed methods include data-level adaptation (DLA), which converts real-world contact F/T to resemble simulation patterns, and feature-level adaptation (FLA), aligning real-world features with simulated ones using knowledge distillation.

% data-level adaptation
For DLA, an MLP-based conversion model trained on a sim-real paired dataset transformed real-world contact F/T to the matched F/T simulation data. The transformation model consisted of two three-layer MLP encoders that used real-world force ($F^R$) and torque ($T^R$) as inputs. The features extracted from each encoder were concatenated and passed through a three-layer MLP for fusion. The fused feature was then utilized to output simulation force and torque through separate three-layer MLP generation heads. The model was trained through the MAE loss with the simulation F/T data ($F^S$ and $T^S$) as the ground truth for sim-to-real adaptation, as illustrated in Fig \ref{fig:figure_5}. 
% feature-level adaptation 
In the FLA, the simulation model was assumed to have the ability to extract essential features for pose estimation that were robust to shapes and poses. The feature extraction capability was transferred to the real-world model using knowledge distillation through MAE loss. The simulation and real-world pose estimation models had the same architecture. The targets of knowledge distillation were force, torque, and fusion features, as shown in Fig.\ref{fig:figure_4}. 
% fine-tuning 
Finally, the model was fine-tuned for the downstream task, using only real-world contact F/T and pose data extracted from sim-real paired datasets.

\section{Experiments}
\label{sec:Experiments}
\subsection{Experimental Environment Setup}

\begin{figure}[h!]
\centering
\includegraphics[width=0.95\linewidth]{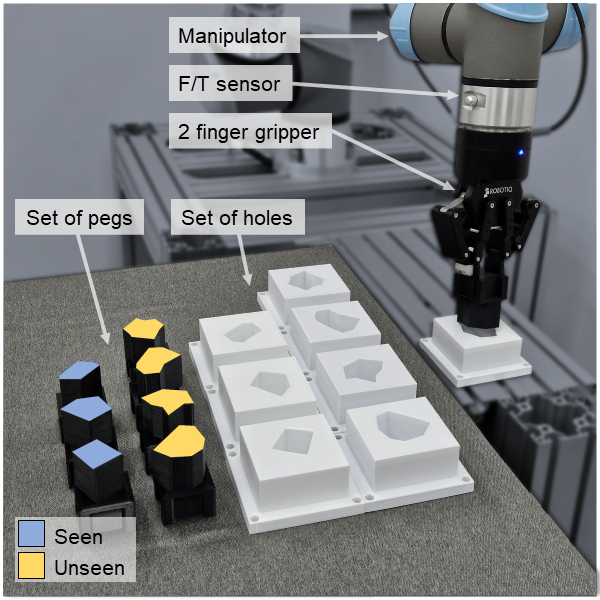}
\caption{Real-world peg-in-hole environment setup.}
\label{fig:figure_6}
% \vspace{-10pt}
\end{figure}

The Isaac Gym simulator was employed as the simulation environment for evaluating the peg-in-hole task, ensuring consistency with the environment utilized for data generation. Position control was implemented to execute the peg-in-hole task following the extrinsic pose estimation. A 6-DoF universal robot (UR5e) equipped with a built-in F/T sensor at the wrist was utilized to capture contact F/T data during task execution and validate the proposed approach in a real-world setting, as shown in Fig. \ref{fig:figure_6}. A specially designed 3D-printed peg with an integrated groove facilitated secure manipulation using the Robotiq 2F-85 parallel gripper, ensuring reliable grasping throughout the peg-in-hole assembly process. In this real-world context, compliance control was implemented for the peg-in-hole task following extrinsic pose estimation that utilized forward dynamic compliance control (FDCC) [35]. The robot system operated at a control frequency of 120 Hz and the F/T sensor sampled data at a rate of 500 Hz. Communication between the control architecture and hardware was facilitated through the ROS (robot operating system) framework.

\subsection{Simulation Evaluation}
{\bf{Extrinsic Pose Estimation Evaluation:}} The model, trained on polygons with 4-6 vertices, was tested on both seen and unseen polygons with 7-10 vertices, using Mean Absolute Error (MAE) for pose error evaluation. The test set consisted of 20 polygons for each n-vertices. For seen polygons, average positional and orientational errors were 1.51 mm and 0.23°, respectively, as listed in Table \ref{tab:table_1}. Despite these averages exceeding the 1 mm peg-hole tolerance, successful assembly is achievable within individual trial tolerances, as detailed following section. Compared to seen polygons, unseen polygons showed slightly higher errors at 1.85 mm for position and 0.44° for orientation. These results indicate the model's effective generalization across different polygonal shapes, maintaining consistent performance for shapes not included in the training.

% ------------------------------------ Fig.4 --------------------------------------

%%%%%%%%%%%%%%%%%%%%%%%%%%%%%%%% figure 7 %%%%%%%%%%%%%%%%%%%%%%%%%%%%%%%%%%%%
\begin{figure*}[ht!]
\centering
\includegraphics[width=\textwidth]{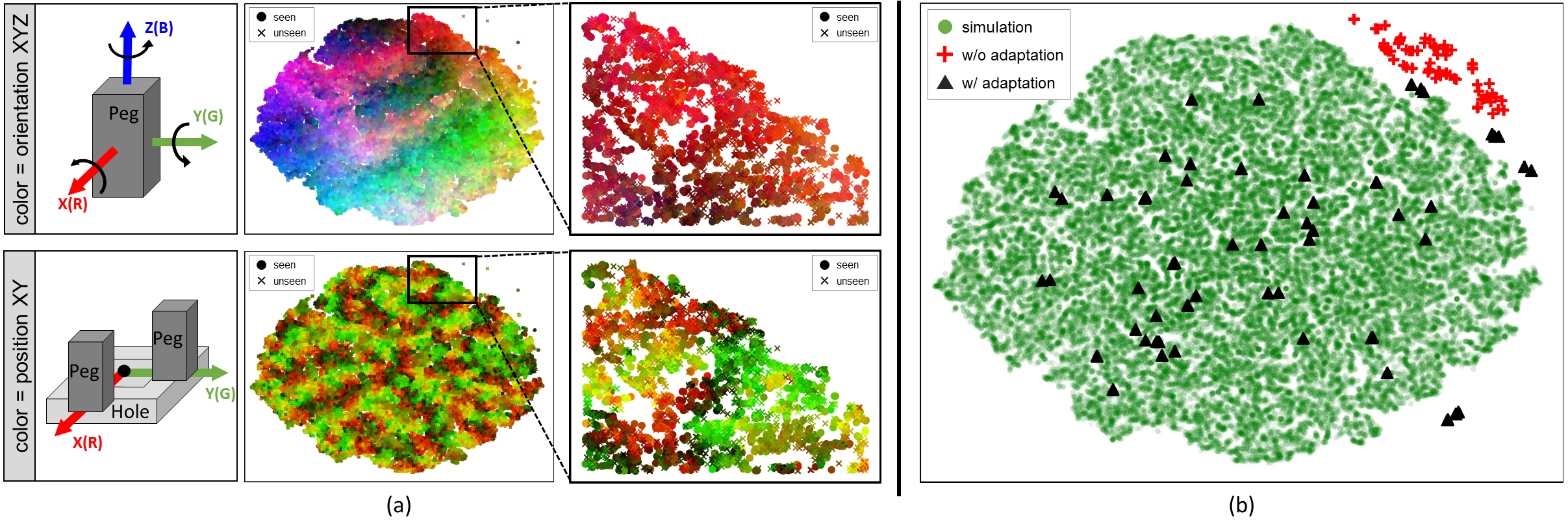}
\caption{t-SNE visualization of feature. (a) t-SNE visualization of simulation data features. Circles and X symbols indicate features of seen and unseen polygons, respectively. The upper images are colored with min-max normalized orientations. Euler X, Y, and Z are mapped to red, green, and blue, respectively. The red and green colors in the lower images are mapped to positions X and Y of the peg, respectively. All t-SNE distributions in the images are the same but colored differently. (b) Visualization of simulation features, where the green circle, red cross, and black triangle represent the sim feature, without sim-to-real, and with sim-to-real, respectively. The t-SNE distribution is the same as the distribution in (a).}
\label{fig:figure_7}
\end{figure*}
%%%%%%%%%%%%%%%%%%%%%%%%%%%%%%%%%%%%%%%%%%%%%%%%%%%%%%%%%%%%%%%%%%%%%%%%%%%%%%%%

%%%%%%%%%%%%%%%%%%%%%%%%%%%%%%%%% Table 1 %%%%%%%%%%%%%%%%%%%%%%%%%%%%%%%%%%%%%
\renewcommand{\arraystretch}{0.8}

% % ------------------------------------ Table 1 --------------------------------------
\begin{table}[h!]
\caption{Evaluation results of extrinsic pose estimation and peg-in-hole.
POS and ORI indicate errors of position XY and orientation XYZ, respectively. SR and AT indicate the success rate, and the average number of trials of peg-in-hole.}
\label{tab:table_1}
\centering
\resizebox{0.46\textwidth}{!}
{
\begin{tabular}{@{}m{3.5em}m{2.5em}cccc@{}}
\toprule
\multicolumn{1}{l}{}    &      & \multicolumn{2}{c}{\vspace{-0.3mm} MAE} & \multicolumn{2}{c}{\vspace{-0.3mm} Peg-in-Hole}  \\ 
\cmidrule(lr){3-4} \cmidrule(lr){5-6} 
                                 & Vertex & POS[mm] & ORI[°] & SR[\%] & AT \\ \midrule
\multirow{4}{*}{\parbox{1cm}{ Seen \newline Polygons}}   & 4    & 1.63    & 0.22    & 96.03   & 2.31   \\
                                 & 5    & 1.51    & 0.23    & 97.50   & 2.36   \\
                                 & 6    & 1.41    & 0.24    & 98.48   & 2.16   \\
                                 & \textbf{Avg}  & \textbf{1.51}    & \textbf{0.23}    & \textbf{97.33}   & \textbf{2.27}   \\ \midrule
\multirow{5}{*}{\parbox{1cm}{ Unseen \newline Polygons}} & 7    & 1.85    & 0.44    & 96.73   & 2.47   \\
                                 & 8    & 1.82    & 0.43    & 95.90   & 2.30   \\
                                 & 9    & 1.82    & 0.45    & 95.98   & 2.39   \\
                                 & 10   & 1.91    & 0.47    & 96.63   & 2.46   \\
                                 & \textbf{Avg}  & \textbf{1.85}    & \textbf{0.44}    & \textbf{96.30}   & \textbf{2.40}   \\ 
\bottomrule
\end{tabular}
}
% \vspace{-10pt}
\end{table}
%%%%%%%%%%%%%%%%%%%%%%%%%%%%%%%%%%%%%%%%%%%%%%%%%%%%%%%%%%%%%%%%%%%%%%%%%%%%%%%%

{\bf{ Extrinsic Pose Estimation Network Feature Analysis:}} To ensure consistent performance on unseen shapes, the deep learning model must be capable of extracting features analogous to those from seen shapes. To verify this ability, the fusion features extracted from the test dataset were visualized using t-SNE in the extrinsic pose estimation model, as shown in Fig. \ref{fig:figure_7}(a). The t-SNE distribution was mapped in red-green-blue (RGB) according to the orientation XYZ of the peg in the upper row and colored in RG according to the position XY in the lower row. The features of the seen and unseen polygons were expressed together and drawn as circles and X markers, respectively.

In the model's feature analysis, orientation distributions, represented in RGB in Fig. \ref{fig:figure_7}(a)'s upper row, display a globally continuous gradation, indicating that similar orientations yield comparable feature values. Contrarily, position distributions, shown in the lower row, present a more discrete coloration, yet closer inspection reveals a local continuity in the gradient. This suggests that orientation predominantly influences the initial phase of position estimation, with subsequent feature extraction being locally modulated based on position. 
In addition, the distribution of the circle and X markers in the enlarged area in Fig.\ref{fig:figure_7}(a) indicates the feature distributions of the seen and unseen polygons. The two markers have similar colors in adjacent areas and are evenly mixed without distinction between markers. This demonstrates that the feature extraction capability was robust to shapes, even those not seen during the training phase.

{\bf{Peg-in-hole Evaluation in Simulation:}}
Initial misalignments were randomly generated in the simulated peg-in-hole evaluation, initiating a multi-point contact strategy for pose estimation and subsequent peg manipulation according to the inferred poses. Each of the 200 test scenarios allowed for a maximum of 5 trials. Successful peg insertion within these trials was recorded as a success, with performance evaluated based on the overall success rate. 
The evaluation resulted in a 97.3\% success rate for seen polygons and 96.3\% for unseen polygons, demonstrating robust performance and shape generalization, as presented in Table \ref{tab:table_1}. Moreover, the system completed tasks in an average of 2.27 and 2.40 trials for seen and unseen polygons, respectively, showcasing consistent performance across both familiar and novel polygon configurations. 

Considering a tolerance of 1 mm, the relationship between the performances of pose estimation and peg-in-hole is shown in Table \ref{tab:table_2}. The table illustrates the maximum values of absolute errors for position and orientation (MA-P, -O) and peg-in-hole results for each trial when evaluating a 5-vertices shape. The initial two trials failed as the pose estimation error exceeded the tolerance. However, in the third trial, the error was reduced to within the tolerance, indicating the success of the peg-in-hole task. As the trials progressed, with a decrease in pose error, the probability of successful insertion increased. Consequently, the proposed framework demonstrated elevated peg-in-hole success rates based on the average trials of 2.27 and 2.40 for seen and unseen polygons.

%%%%%%%%%%%%%%%%%%%%%%%%%%%%%%%% TABLE 2 %%%%%%%%%%%%%%%%%%%%%%%%%%%%%%%%%%%%%%%%%%%%%%%%
\begin{table}[h!]
{\small
\caption{Example of peg-in-hole evaluation on simulation.
MA-P and MA-O indicate the maximum value of absolute position error XY and orientation error XYZ. The bottom row visualizes the results of peg-in-hole by moving the peg to the estimated pose.}
\label{tab:table_2}
\centering
\resizebox{0.47 \textwidth}{!}
{
% \begin{tabularx}{0.45\textwidth}{@{} *{6}{X} @{}}
% \begin{tabularx}{0.45\textwidth}{@{} *{6}{X} @{}}
\begin{tabular}{@{}lccccc@{}}
\toprule
          & Trial 1             & Trial 2            & Trial 3                 & Trial 4  & Trial 5  \\ \midrule
MA-P [mm]       & 1.46          & 1.67         & \textbf{0.91}              & -            & -  \\
MA-O [°]   & 1.63          & 0.47         & \textbf{0.49}             & -            & -  \\
S/F            & Fail          & Fail         & \textbf{Success}            & -            & -  \\ \bottomrule
\noalign{\smallskip} 
\multicolumn{3}{c}{Result of Trials 1 (Fail)} & \multicolumn{3}{c}{Result of Trial 3 (Success)} \\
\multicolumn{3}{c}{\includegraphics[width=0.4\linewidth]{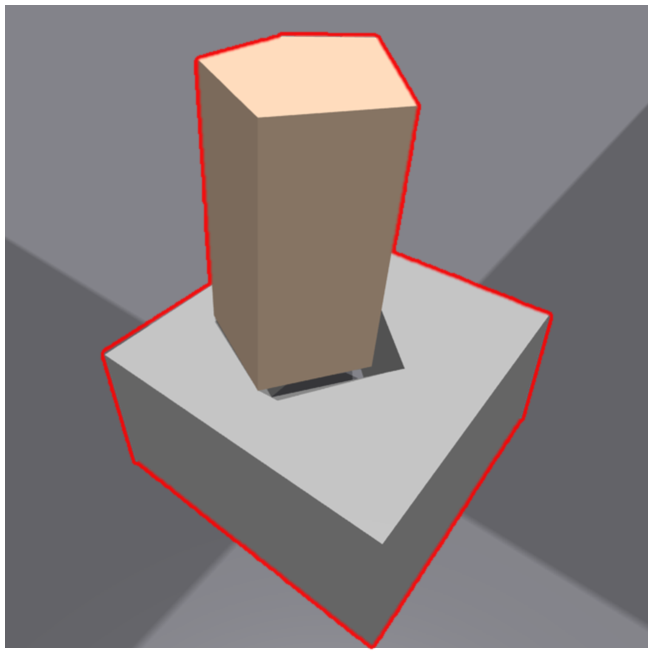}}
& \multicolumn{3}{c}{\includegraphics[width=0.4\linewidth]{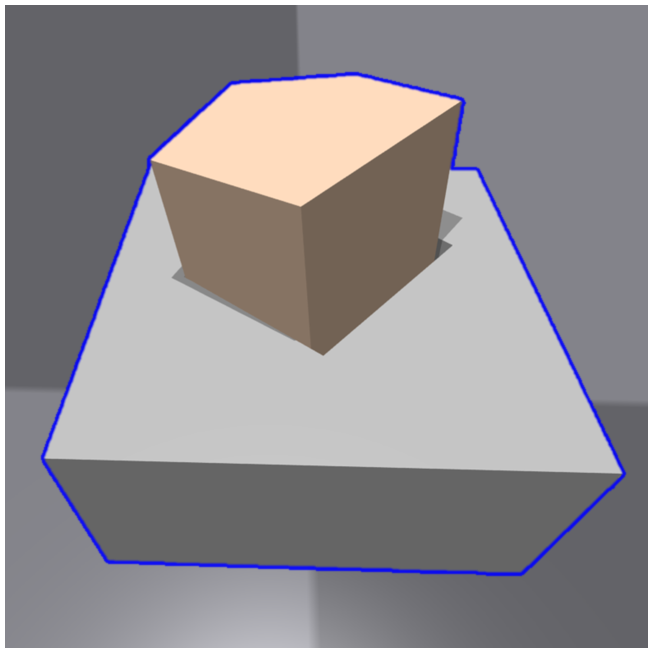}} \\ \bottomrule
\end{tabular}
}
}
% \vspace{-10pt}
\end{table}
%%%%%%%%%%%%%%%%%%%%%%%%%%%%%%%% TABLE 2 %%%%%%%%%%%%%%%%%%%%%%%%%%%%%%%%%%%%%%%%%%%%%%%%

\subsection{Real-world Adaptation}
{\bf{Sim-Real Paired Dataset:}}
In the real-world setting, polygons from the simulation dataset were 3D printed and utilized to generate a paired sim-real dataset. Similar to the extrinsic pose estimation model training set, only polygons with 4 to 6 vertices were used for sim-to-real adaptation. Misalignment data for peg-hole interactions in 40 random poses for each 4 to 6 vertices polygon were gathered in both simulation and real-world scenarios. Additionally, data from 20 poses, involving both "seen" and "unseen" polygons, were collected for evaluating pose estimation in real-world scenarios. The process of collecting real-world data for each shape took approximately 10 min.

\begin{figure}[h!]
\centering
\includegraphics[width=\linewidth]{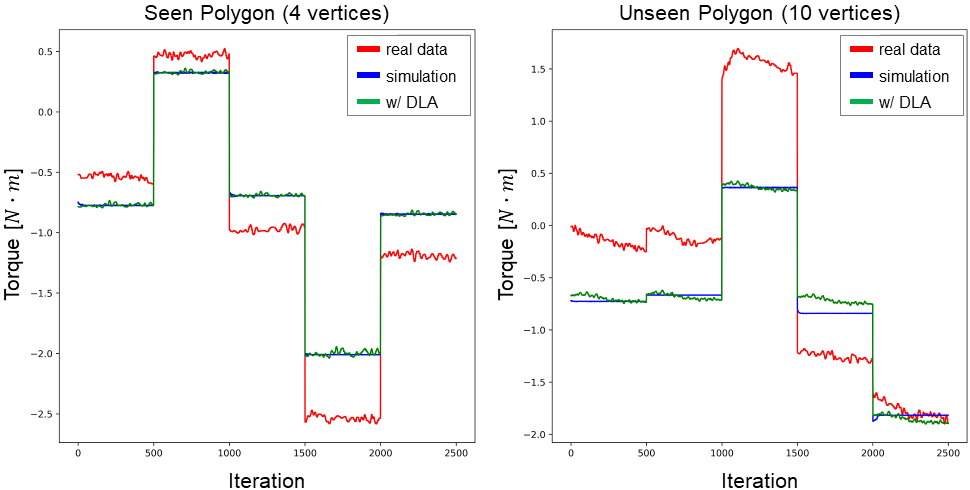}
\caption{Examples of data-level adaptation. DLA indicates the proposed method of data-level adaptation.}
% \vspace{-10pt}
\label{fig:figure_8}
\end{figure}

{\bf{Extrinsic Pose Estimation in Real World:}} 
For DLA, real-world contact F/T was converted to their corresponding simulation counterparts using an MLP-based network. Torque data examples from real-world, simulation, and converted with DLA, are illustrated in Fig. \ref{fig:figure_8} with red, green, and blue lines. In both seen and unseen polygons, the converted torque data exhibited a similar pattern to that observed in the simulation. In FLA, features extracted from simulation data, real-world data without adaptation, and real-world data after sim-to-real adaptation through knowledge distillation were visualized collectively using t-SNE. The red cross in Fig. \ref{fig:figure_7}(b) indicates a separate distribution from the green circle and black triangle. However, the black triangles shifted the distribution among the green circles, suggesting that real-world features aligned with simulation features after knowledge distillation. 

The extrinsic pose estimation performance was evaluated using a collected real-world dataset after sim-to-real adaptation. Significant pose errors, exceeding 5 mm in position and 10° in orientation, occurred when applying a pose estimation model trained solely on simulation data to a real environment without any adaptation process, as shown in Table \ref{tab:table_3}. Each individual method, when reported separately, exhibited an improved pose estimation performance compared to the \textit{w/o adaptation}. The proposed comprehensive methodology that incorporated all data achieved the best pose estimation performance, indicating efficient use of the collected sim-real paired dataset.

% ------------------------------- Fig.7--------------------------------------

%%%%%%%%%%%%%%%%%%%%%%%%%%%%%%%% TABLE 3 %%%%%%%%%%%%%%%%%%%%%%%%%%%%%%%%%%%%%%%%%%%%%%%%
\begin{table}[h]
\caption{Pose estimation errors of sim-to-real adaptation methods. POS and ORI indicate errors of position XY and orientation XYZ, respectively.}
\label{tab:table_3}
\centering
\resizebox{0.47\textwidth}{!}
{
\begin{tabular}{@{}lcccc@{}}
\toprule
                & \multicolumn{2}{c}{Seen Polygons} & \multicolumn{2}{c}{Unseen Polygons} \\ 
                \cmidrule(lr){2-3}                \cmidrule(lr){4-5}
Method          & POS[mm]  & ORI[°]  & POS[mm] & ORI[°]        \\ \midrule
w/o Adaptation  & 5.37       & 14.63     & 5.16       & 10.86      \\
DLA             & 1.73       & 0.46      & 4.39       & 1.88       \\
FLA             & 2.07       & 0.79      & 5.12       & \textbf{1.77} \\
Fine-tuning     & 4.77       & 0.39      & 5.91       & 2.03       \\
\textbf{Ours}   & \textbf{0.48} & \textbf{0.17} & \textbf{3.91} & 1.85 \\
\bottomrule
\end{tabular}
}
\end{table}
%%%%%%%%%%%%%%%%%%%%%%%%%%%%%%%% TABLE 3 %%%%%%%%%%%%%%%%%%%%%%%%%%%%%%%%%%%%%%%%%%%%%%%%

{\bf{Peg-in-hole Evaluation in Real World:}}
The peg-in-hole tasks were evaluated in a real-world environment using the proposed sim-to-real adapted pose estimation model. Each shape was evaluated 20 times using the same protocol as that in the simulation. 
Polygons of 4 to 6 vertices were classified as seen shapes and utilized for domain adaptation, while those with 7 to 10 vertices were designated as unseen, as shown in Fig. \ref{fig:figure_6}. Comparative benchmarks encompassed three methods: a standard spiral-search method, PolyFit without adaptation that was solely dependent on linear scaling between simulated and real data, and PolyFit with sim-to-real adaptation. 
The results for the adapted model demonstrated a substantial enhancement in performance, evidenced by success rates of 86.6\% and 85.0\% for seen and unseen polygons, respectively, as listed in Table \ref{tab:table_4}. 
The success rates achieved by PolyFit with adaptation markedly exceeded those obtained without adaptation, 38.3\% for seen shapes and 28.8\% for unseen shapes. Additionally, these figures represent a substantial improvement over the success rates of 18.3\% for seen shapes and 17.5\% for unseen shapes achieved by the spiral search method. This notable difference emphasized the capability of the proposed pose estimation model to accommodate unseen polygon shapes and validated the successful transition and adaptability of the simulation-trained model to real-world challenges.

%%%%%%%%%%%%%%%%%%%%%%%%%%%%%%% Table 4 %%%%%%%%%%%%%%%%%%%%%%%%%%%%%%%%%%%%%%%%%%%%%%%%%
\renewcommand{\arraystretch}{1.1}
\begin{table*}[t]
\caption{Evaluation results of peg-in-hole in the real-world environment}
\label{tab:table_4}
\large
\centering
\resizebox{\textwidth}{!}
{
{
\begin{tabular}{@{}lcccccccc@{}}
\toprule
 & \multicolumn{3}{c}{Success rate of seen polygons [\%]} & \multicolumn{4}{c}{Success rate of s unseen polygons [\%]} & \multicolumn{1}{c}{Average Trials} \\
\cmidrule(lr){2-4} \cmidrule(lr){5-8} \cmidrule(lr){9-9}
 & \multicolumn{1}{c}{\includegraphics[height=6em]{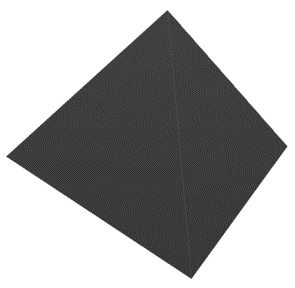}}  & \multicolumn{1}{c}{\includegraphics[height=6em]{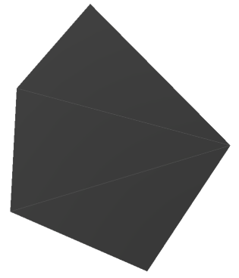}}  & \multicolumn{1}{c}{\includegraphics[height=5.8em]{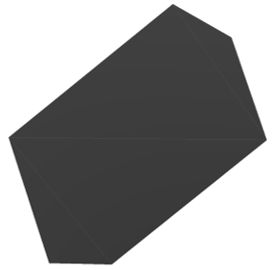}}  & \multicolumn{1}{c}{\includegraphics[height=5.8em]{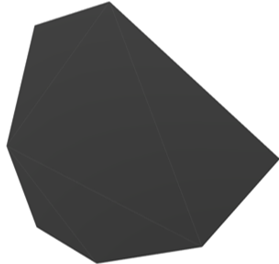}}  & \multicolumn{1}{c}{\includegraphics[height=6em]{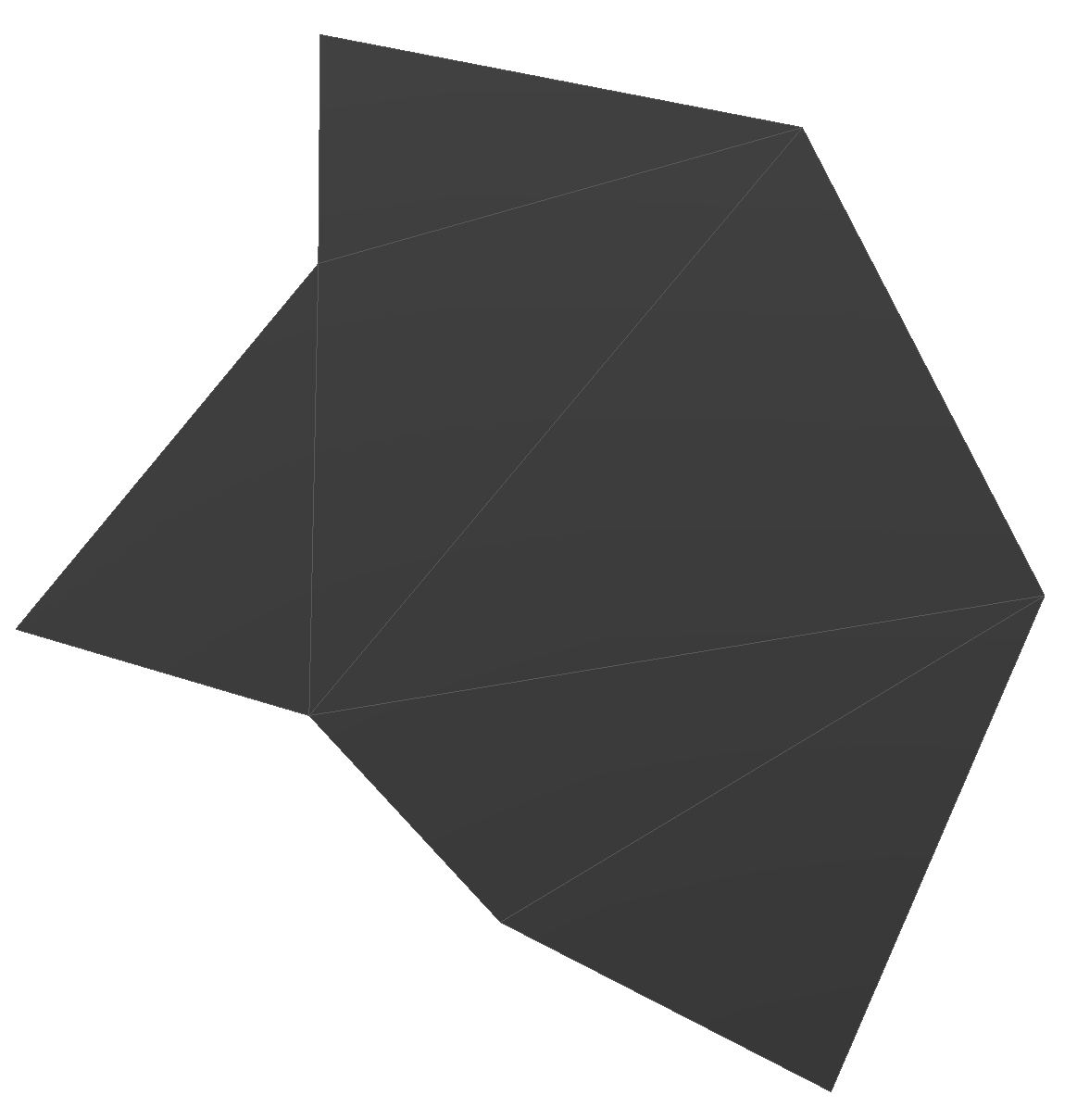}}  & \multicolumn{1}{c}{\includegraphics[height=5.8em]{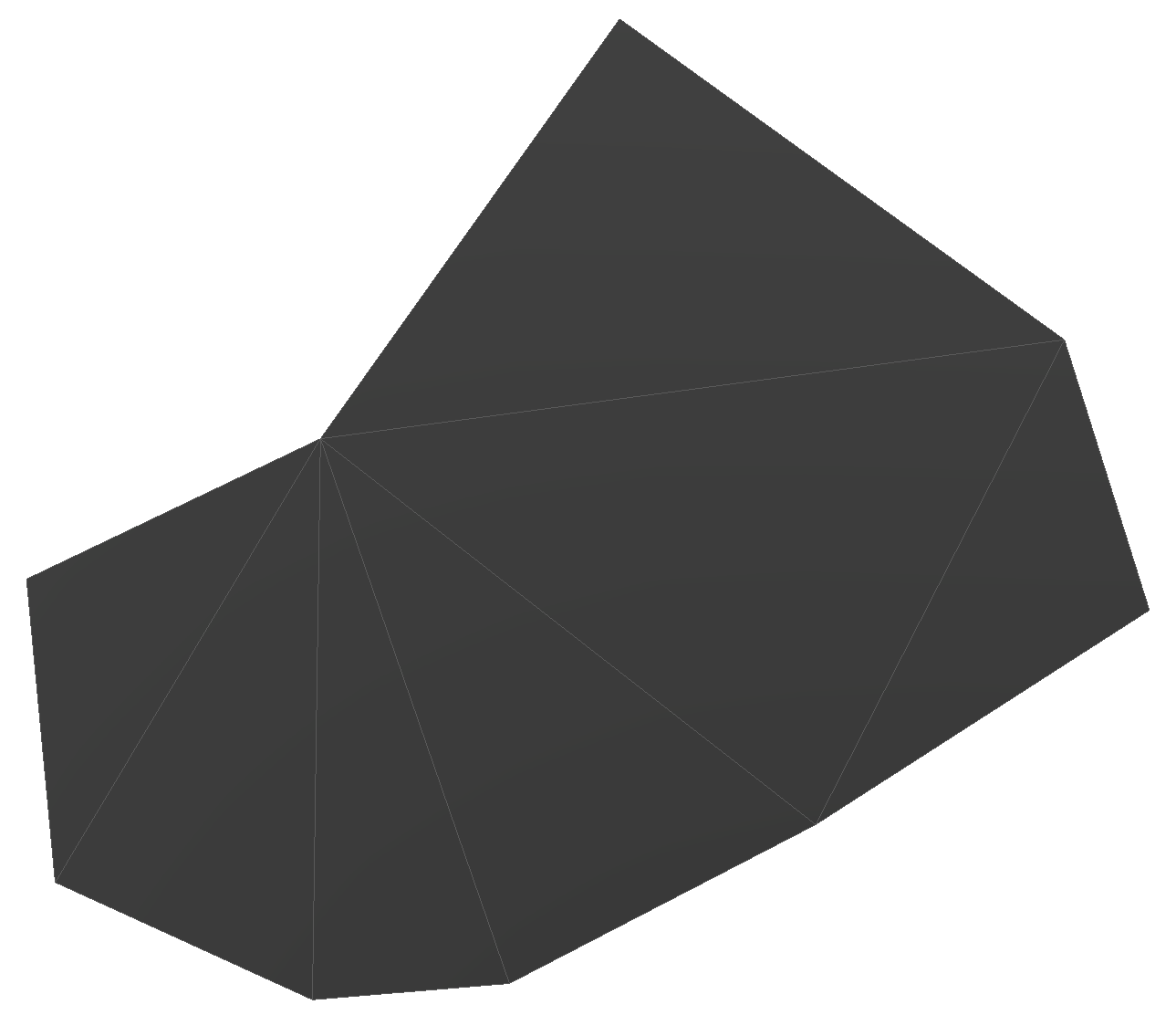}}  & \multicolumn{1}{c}{\includegraphics[height=6.1em]{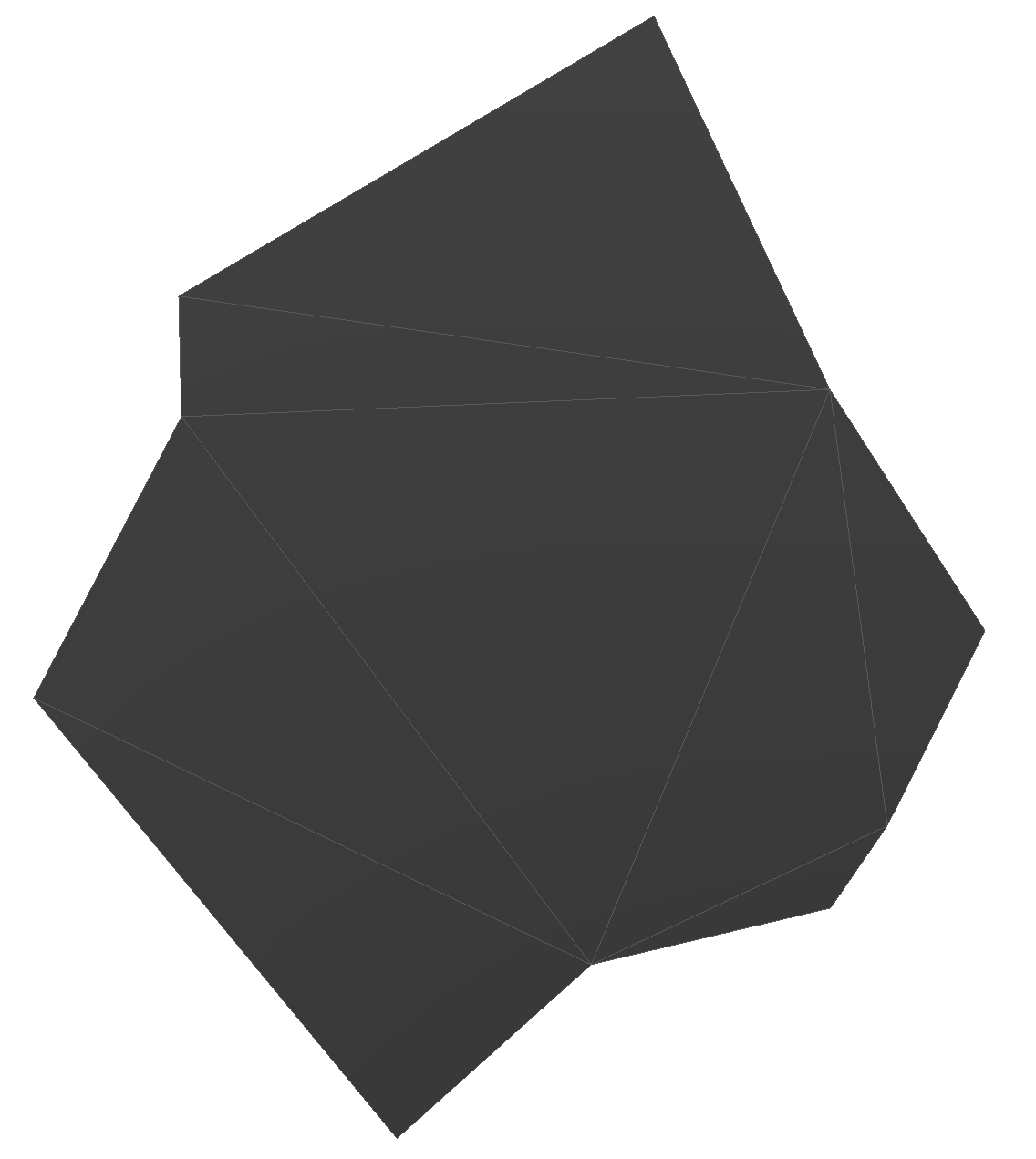}} & \\
Method & 4-vertices & 5-vertices & 6-vertices & 7-vertices & 8-vertices & 9-vertices & 10-vertices & \\
\midrule
Spiral Search & 15.0 (3/20) & 20.0 (4/20) & 20.0 (4/20) & 25.0 (5/20) & 20.0 (4/20) & 20.0 (4/20) & 15.0 (3/20) & - \\
PolyFit (w/o AD) & 30.0 (6/20) & 40.0 (8/20) & 45.0 (9/20) & 35.0 (7/20) & 30.0 (6/20) & 20.0 (4/20) & 30.0 (6/20) & 3.11 \\
% PolyFit (Fine-tuning) & - & - & - & - & - & - & - & - \\
\textbf{PolyFit (w/ AD)} & \textbf{85.0 (17/20)} & \textbf{85.0 (17/20)} & \textbf{90.0 (18/20)}& \textbf{95.0 (19/20)} & \textbf{80.0 (16/20)} & \textbf{75.0 (15/20)} & \textbf{90.0 (18/20)} & \textbf{2.27} \\
\bottomrule
\end{tabular}
    }
    }    
\end{table*}
% %%%%%%%%%%%%%%%%%%%%%%%%%%%%%%%%%%%%%%%%%%%%%%%%%%%%%%%%%%%%%%%%%%%%%%%%%%%%%%%%

\section{Conclusions}
\label{sec:conclusion}

This study introduces an F/T-based framework capable of executing 5-DoF peg-in-hole assembly, effectively addressing key challenges in robotic assembly tasks. The extrinsic pose estimation model exhibits noteworthy performance on both seen and unseen polygonal shapes, achieved through extensive simulation-based training. The introduction of sim-to-real adaptation, leveraging a sim-real paired dataset, proves instrumental in maintaining high-performance levels with unseen polygons. This underscores the framework's versatility across diverse geometries and operational environments. The findings of this research have the potential to mitigate inefficiencies and safety concerns associated with real-world data acquisition in the future. By utilizing a dynamic simulator for learning and data collection, the framework ensures robust performance across various component geometries and conditions, eliminating the need for extensive real-world data collection. Future work will focus on incorporating a closed-loop framework to enhance robustness and extend the application to more practical assemblies, such as connectors and cables.

\section*{Acknowledgments}
\begin{spacing}{0.4}
{\scriptsize This work was supported by Electronics and Telecommunications Research Institute (ETRI) grant funded by the Korean government. (23ZR1100, A Study of Hyper-Connected Thinking Internet Technology by autonomous connecting, controlling and evolving ways). And also supported by Artificial intelligence industrial convergence cluster development project funded by the Ministry of Science and ICT(MSIT, Korea)\&Gwangju Metropolitan City.}
\end{spacing}

\bibliography{bibtex/refer.bib}
\bibliographystyle{IEEEtran}

\end{document}